# Optical Engineering

# Automatic trajectory measurement of large numbers of crowded objects

Hui Li
Ye Liu
Yan Qiu Chen

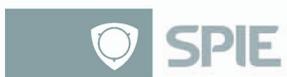

# Automatic trajectory measurement of large numbers of crowded objects

**Hui Li**
**Ye Liu**
**Yan Qiu Chen**
Fudan University
School of Computer Science
825, Zhangheng Road-1159, Cailun Road
Shanghai, Shanghai 201203, China
E-mail: hui_li@fudan.edu.cn

**Abstract.** Complex motion patterns of natural systems, such as fish schools, bird flocks, and cell groups, have attracted great attention from scientists for years. Trajectory measurement of individuals is vital for quantitative and high-throughput study of their collective behaviors. However, such data are rare mainly due to the challenges of detection and tracking of large numbers of objects with similar visual features and frequent occlusions. We present an automatic and effective framework to measure trajectories of large numbers of crowded oval-shaped objects, such as fish and cells. We first use a novel dual ellipse locator to detect the coarse position of each individual and then propose a variance minimization active contour method to obtain the optimal segmentation results. For tracking, cost matrix of assignment between consecutive frames is trainable via a random forest classifier with many spatial, texture, and shape features. The optimal trajectories are found for the whole image sequence by solving two linear assignment problems. We evaluate the proposed method on many challenging data sets. © *2013 Society of Photo-Optical Instrumentation Engineers (SPIE)* [DOI: 10.1117/1.OE.52.6.067003]

Subject terms: trajectory measurement; variance minimization active contour; dual ellipse locator; trainable assignment cost matrix; linear assignment problem.

Paper 130298 received Feb. 25, 2013; revised manuscript received May 10, 2013; accepted for publication May 15, 2013; published online Jun. 18, 2013.

## 1 Introduction

Complex dynamics displayed by natural systems, such as fish schools, bird flocks, and cell groups, often demonstrate fascinating phenomenon, e.g., coherent structures of swimming micro-organisms[1] and have attracted great attention from scientists for years.[2] Although many numerical models[3,4] have been proposed to simulate such phenomena, theoretical assumptions behind them without the support of credible trajectory are insufficient to reveal the underlying mechanisms of these fascinating phenomena.[5] Therefore, trajectory measurement of each individual is vital for quantitative and high-throughput study of the collective behavior.

Manual tracking is, however, tedious, time consuming, and error prone especially when measuring trajectories of large numbers of objects during a long sequence. It is, therefore, highly desirable to develop an automatic and robust tracking system. We mainly focus on an automatic trajectory measurement of oval-shaped objects because in the natural environment, many objects can be simplified as a solid ellipse, such as fish, cells, the human head, birds, drosophila, and mitochondria in microscopic imagery. Our motivation in this work is the task of detecting mitochondria and biological cells.[6] The similar assumption was often made by many other researchers such as Refs. 7–9.

For an automatic tracking system, the first step is object detection and segmentation. Recently, many researchers[9–11] have contributed to the detection and segmentation of oval-shaped objects. However, when dealing with large numbers of crowded objects, as shown in Fig. 1, current methods are prone to failure because of blurred boundaries and frequent occlusions. Some researchers have attempted to overcome these difficulties.[7] Some proposed a difference of Gaussian model to detect a large number of moving cells, but the data association with frequent occlusions is unclear.[8] These studies had retrieved hundreds of cells under phase contrast microscopy, but they simplified each individual as a point, which loses the important shape and texture information.

As an important research area, trajectory measurement (tracking) in video sequence has developed over several years. There are many useful tracking methods, including the nearest neighbor algorithm (NN), the joint likelihood filter (JLF), the joint probabilistic data association (JPDA) algorithm, and the multiple hypothesis filter (MHF). Unfortunately, JLF, JPDA, and MHF attempt to solve an nondeterministic polynomial-time hard data association problem. Unless the number of objects is small and track crossings are not common, these methods are often impractical to employ in practice. Rasmussen and Hager[12] and Cox and Miller[13] extended the original JLF, JPDA, and MHF algorithms to track objects by complex feature combinations. Recently, tracking has been modeled as some kind of first-order Markov chain such as Kalman filter,[5,14] extended Kalman filter,[15] and particle filter.[16] However, all of these methods need to define the association cost between two objects. To the best of our knowledge, most of the current methods are prone to using spatial and texture features but ignoring the important shape feature, which is very useful in distinguishing large numbers of deformable objects, such as fish and cells. In addition, the cost matrix of data assignment of current methods between consecutive frames was usually defined by an energy function which fused many feature terms, such as spatial distance and texture similarity. However, this kind of methods has an obvious weakness in that the cost matrix highly depends on the weight of each term and a suitable normalization method should be employed for each term separately, which is very subjective.

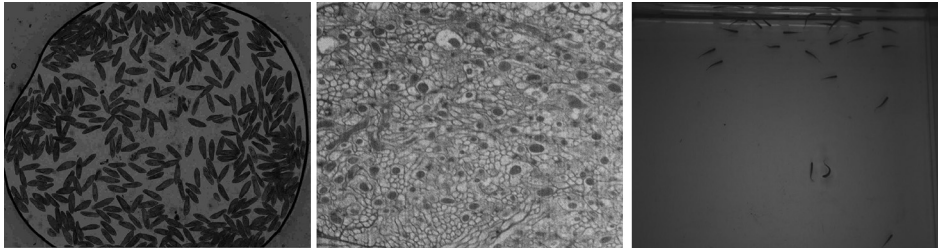

**Fig. 1** Detection and tracking of large numbers of objects with frequent occlusions. From left to right: paramecium group, mitochondrion, and zebra fish school.

To overcome the limitation of current methods, we present an automatic and effective framework to detect and track large numbers of crowded oval-shaped objects. We first prove an interesting dual ellipse inference (DEI) and use a novel DEI-based locator to obtain the coarse position of visually similar targets and then propose a variance minimization active contour model (VM-ACM) to get the optimal segmentation results iteratively. For data association, the costs are trainable on positive assignment samples with fully using spatial–temporal information, texture, and shape features. An optimal set of assignments is found for the whole image sequence by solving two steps of linear assignment problems (LAP). The proposed framework works effectively on many challenging data sets and has a good ability to overcome frequent occlusions. Experiment results show that the proposed framework outperforms the state-of-the-art methods. Our contributions are listed as follows:

- We propose an interesting inference that the zero and minimum level sets of arbitrary difference of two Gaussian (DoG) functions of the same center, phase, and maximum are circumscribed and inscribed ellipses of a rectangle. We term this inference as DEI. This inference is very useful because with the property of DEI, it is very easy for users to transform the evolution of DoG model to the evolution of the rectangle. In addition, this inference reveals an important relationship of its four variances as formulated in Eq. (4). We can calculate these four variances only with three parameters. This inference can be used in many other applications, such as scale invariant feature transform (SIFT)[17] and band-pass filter.
- Trajectory measurement for large numbers of a high-density population is a challenging work. In this paper, we give two useful clues: clue-1 in Sec. 2.2 and clue-2 in Sec. 2.4. We think that these two clues give a possible approach to overcome the challenges in detection and tracking of crowded objects.
- We propose a novel ACM based on VM energy function to get the fine segmentation of each detected object because fine segmentation results can not only be used to measure the shape of different individuals but also provide precise data for us to extract features of trainable assignment costs. This method can also be used to segment many other objects separately.
- As opposed to current methods, we do not define a subjective function to measure the similarity of two objects. In contrast, a trainable cost matrix is given via random forest (RF) classifier with many spatial–temporal, texture, and shape features. Experiment results show that this kind of method is more reliable than the state-of-the-art methods.
- We formulate the tracking problem as two steps of LAP, which can handle frequent occlusions and obtain complete trajectories of high-density population.

## 2 Detection and Segmentation

Detection and segmentation for large numbers of a high-density population are a very challenging problem because of visually similar features and frequent occlusions. In this section, we first prove an interesting DEI and then present a novel DEI-based dual ellipse locator (DEL) to detect the coarse positions of objects. At last, we propose a variance minimization active contour method (VM-ACM) with a novel energy function to adopt the fine segmentation result for each target. Specifically, the proved DEI and VM-ACM can also be used in many other applications, which will be discussed in Sec. 5.

### 2.1 Dual Ellipse Inference

In this section, we will prove an interesting property that the zero and minimum level sets of arbitrary DoG functions of the same center, phase, and maximum are circumscribed and inscribed ellipses of a rectangle. We call this property DEI.

Let $f(x,y) = G_2 - G_1$ be the DoG functions $G_1$ and $G_2$ of the same center, phase, and maximum. Let $\sigma_{x_i}$ and $\sigma_{y_i}$ be the variances of $G_i$. To simplify the discussion, we use normal Gaussian functions in the remainder of the section.

#### 2.1.1 Zero level set

To obtain the zero level set of $f(x,y)$, let $f(x,y) = 0$, then we have Eq. (1):

$$\left(\frac{x^2}{2\sigma_{x_2}^2} + \frac{y^2}{2\sigma_{y_2}^2}\right) - \left(\frac{x^2}{2\sigma_{x_1}^2} + \frac{y^2}{2\sigma_{y_1}^2}\right) = \ln\left(\frac{\sigma_{x_1}\sigma_{y_1}}{\sigma_{x_2}\sigma_{y_2}}\right). \qquad (1)$$

In two dimensions, the level sets of $f(x,y)$ will always be ellipses. Let the standard format of the zero level set Eq. (1) be $x^2/a^2 + y^2/b^2 = 1$, we then can obtain

$$\begin{cases} \frac{1}{2\sigma_{x_2}^2} - \frac{1}{2\sigma_{x_1}^2} = \frac{1}{a^2} \ln \frac{\sigma_{x_1}\sigma_{y_1}}{\sigma_{x_2}\sigma_{y_2}} \\ \frac{1}{2\sigma_{y_2}^2} - \frac{1}{2\sigma_{y_1}^2} = \frac{1}{b^2} \ln \frac{\sigma_{x_1}\sigma_{y_1}}{\sigma_{x_2}\sigma_{y_2}} \end{cases}. \quad (2)$$

### 2.1.2 Minimum level set

To get the minimum level set of $f(x,y)$, we calculate its first derivatives on $x$ and $y$, then we will have $G_1/\sigma_{x_1}^2 = G_2/\sigma_{x_2}^2$ and $G_1/\sigma_{y_1}^2 = G_2/\sigma_{y_2}^2$. Let $G_1/G_2 = k^2$ and the standard format of the minimum level set be $x^2/c^2 + y^2/d^2 = 1$. It is easy to verify Eq. (3):

$$\begin{cases} \frac{1}{2\sigma_{x_2}^2} - \frac{1}{2\sigma_{x_1}^2} = \frac{1}{c^2} \ln \frac{\sigma_{x_1}^3\sigma_{y_1}}{\sigma_{x_2}^3\sigma_{y_2}} \\ \frac{1}{2\sigma_{y_2}^2} - \frac{1}{2\sigma_{y_1}^2} = \frac{1}{d^2} \ln \frac{\sigma_{x_1}^3\sigma_{y_1}}{\sigma_{x_2}^3\sigma_{y_2}} \end{cases}. \quad (3)$$

Now, from Eqs. (2) and (3), we can infer several conclusions as Eq. (4):

$$\begin{cases} c = \sqrt{2}a, & d = \sqrt{2}b \\ \sigma_{x_2} = \sqrt{\frac{(k^2-1)a^2}{4k^2 \ln k}}, & \sigma_{y_2} = \sqrt{\frac{(k^2-1)b^2}{4k^2 \ln k}} \\ \sigma_{x_1} = \sqrt{\frac{(k^2-1)a^2}{4 \ln k}}, & \sigma_{y_1} = \sqrt{\frac{(k^2-1)b^2}{4 \ln k}} \end{cases}. \quad (4)$$

From Eq. (4), we can obtain an interesting property that the zero and minimum level sets of arbitrary $f(x,y)$ are circumscribed and inscribed ellipses of a rectangle with width $a$ and height $b$. The inference DEI holds true.

We should use four parameters ($\sigma_{x_1}$, $\sigma_{y_1}$, $\sigma_{x_2}$, and $\sigma_{y_2}$) to control evolution of $f(x,y)$ without DEI. But now, we only need three parameters [($a$, $b$, and $k$ in Eq. (4)]. That is, DEI can reduce the dimensions of DoG model without loss of precision. This property is very useful and can be used in many applications as discussed in Sec. 5. Let $S$ be the rectangle and $S_1$ be the inscribed ellipse, then $S_2$ will be $S - S_1$. DEI has many other good properties: (1) let $p = (x,y)$ denote a point in $S$, we have $f(p) > 0$ if $p \in S_1$ and $f(p) < 0$ if $p \in S_2$; (2) let $o = (x,y)$ denote the center of the rectangle and $\text{dist}(o,p)$ denote the Euclidean distance between $o$ and $p$. We have $f(p_1) > f(p_2)$ if $\text{dist}(o,p_1) < \text{dist}(o,p_2)$.

### 2.2 Coarse Detection via DEI

One can see that crowded objects such as cells are oval-shaped gray blobs of similar and yet slightly varied size and shape[18] as illustrated in Fig. 1. A large range of connected objects is difficult to distinguish because of frequent occlusions and distractions. From a raw image, we find a clue that there are about 2 to 4 blank corner regions at the head and tail of most of the objects even in high-density populations, which is very like our DEI model. We term this clue as clue-1. Similar clues also occurs in zebra fish and mitochondria data sets.

Based on the above clue-1 and the properties of DEI (Sec. 2.1), we propose an adaptive DEL to fit the coarse positions of crowded oval-shaped objects. DEL is a rectangular model $S$ (the width is $a$ and the height is $b$). Let $T(x,y)$ denote the normalized gray value of point $(x,y)$ in raw image $T$. We define the fit value as

$$\gamma = \iint_{(x,y)\in S} f(x,y) * T(x,y) \mathrm{d}x \mathrm{d}y. \quad (5)$$

For an individual in $T$, there will be a region of the same size with $S$, who could fit an appropriate scale of DEL. The fit value depends on how much the object looks like an oval. Based on this, we transform the detection problem to finding regions whose $\gamma$ is larger than threshold $\lambda$.

Since different targets may have slight size changes, we use an adaptive method to regulate $\sigma_{x_i}$ and $\sigma_{y_i}$ of $f(x,y)$. Generally, these four parameters ($\sigma_{x_1}$, $\sigma_{y_1}$, $\sigma_{x_2}$, and $\sigma_{y_2}$) should be specified by users,[7] yet with the proved DEI (see details in Sec. 2.1), we can transform this problem to changing width $a$ and height $b$ of the rectangle as well as the parameter $k$ [see Eq. (4)]. Let $c$ be the rangeability of $a$ and $b$. Adaptive parameters $k$, $a$, $b$, and $c$ are used to control the scalability and evolution of DEL because different individuals may have different sizes. Although there are many parameters in this section, it is very easy for users to specify suitable parameters for different data set, which will be discussed in Sec. 4.2.

In one region, the DEL may detect more than one individual whose fit values are larger than $\lambda$. We use the overlapping area to judge whether the locator repeats the same object. In our experiment, we prefer to select the target whose position is of larger fit value as a better target.

### 2.3 Variance Minimization Active Contour Model

Accurate segmentation results and object shape are very important to biologists.[8,18] Feature extraction based on the segmentation results rather than the detection results is more precise, which will be used in object tracking and linking steps as discussed in Sec. 3. When addressing segmentation, active contour without edge method[19] with energy function Eq. (6) is often used in computer community.

$$F(C) = \mu \cdot \text{Length}(C) + \nu \cdot \text{Area}(C_{\text{in}}) + \lambda_1$$
$$\cdot \int_{C_{\text{in}}} |T(p) - c_1|^2 \mathrm{d}p + \lambda_2 \cdot \int_{C_{\text{out}}} |T(p) - c_2|^2 \mathrm{d}p \quad (6)$$

where $C$ denotes the evolving curve in $\Omega \in R^2$, as the boundary of an open subset $\omega$ of $\Omega$, i.e., $\omega \in \Omega$ and $\omega = \partial\Omega$. Let $C_{\text{in}}$ denote the region $\omega$ (object) and $C_{\text{out}}$ denote the region $\Omega - \bar{\omega}$ (background). $C_{\text{in}} \bigcup C_{\text{out}} = \Omega - C$. $T(p)$ denotes the gray value of pixel $p = (x,y)$ in raw image $T$. $c_1$ and $c_2$ are mean intensity inside and outside of curve $C$, which is represented implicitly via a Lipschitz function $\phi$ by $C = \{p|\varphi(p) = 0\}$ and the evolution of which is given by the zero-level curve at time $t$ of the function $\phi(t,p)$. Specifically, we have $C_{\text{in}} = \{p|\phi(t,p) < 0\}$ and $C_{\text{out}} = \{p|\phi(t,p) > 0\}$.

In Chan–Vese's energy function, Eq. (6), we find that with the evolution of curve $\phi$, item Length($C$) will be smaller and smaller. In the case of Fig. 2, however, we find a counterexample in that the length of curve $\phi$ is longer than the object's boundary, i.e., Length($C$) is not always decreasing with the evolution of $\phi$. Therefore, Eq. (6) will not obtain the optimal segmentation when dealing with objects with a concave surface.

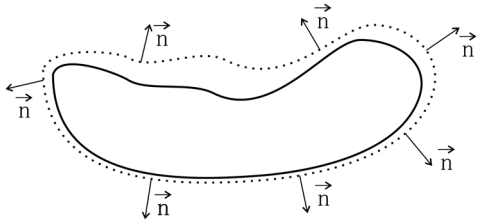

**Fig. 2** Chan–Vese's active contour without edges. The solid line denotes the object boundary. The dotted line denotes the active contour $\varphi$ in time $t$. $\vec{n}$ is the exterior normal vector to the contour.

To overcome their shortcomings, we propose a novel VM-ACM to segment each detected individual. Specifically, in the discussion of this section, each coarsely detected object can be seen as an image patch, which is processed by VM-ACM separately. We formulate VM-ACM as

$$F(\varphi) = \lambda_1 \text{Var}(C_{\text{in}}) + \lambda_2 \text{Var}(C_{\text{out}}) - \lambda_3 \text{Var}(C), \quad (7)$$

where $\text{Var}(\cdot)$ denotes intensity variance of the corresponding region. Different from the Chan–Vese's method,[19] our segmentation results could guarantee VM, and the parameter $\lambda_i$ rarely relies on experimental objects because each item in Eq. (7) is normalized. Since the number of points in $C$ is far less than $C_{\text{in}} \bigcup C_{\text{out}}$, we can assume that the third item of Eq. (7) is constant during the whole evolution.

Let $H(\phi)$ be Heaviside function, $\delta(\phi)$ be one-dimensional Dirac measure, and $h(p)$ be probability density function, which is used to describe the normalized weight of each pixel. We assume that $h(p)$ is $f(x, y)$ of its corresponding DEL. To obtain a global minimizer, independent of the position of the initial curve, we borrow the idea in Ref. 19 and use the regularization of $H(\phi)$ as $H(\varphi, \epsilon) = \frac{1}{2}(1 + \frac{2}{\pi}\arctan(\frac{\varphi}{\epsilon}))$. Generally, $\epsilon$ in $H(\phi)$ is specified as 1.

Let $F(\varphi) = \int g(\varphi(t, p)) dp$ and $\phi(p, 0)$ be the initial contour. After minimizing $F(\phi)$ with the gradient projection method, we can obtain

$$\frac{\partial \varphi}{\partial t} = -\lambda_1 T(p)^2 h(p_1) \delta(\varphi) + \lambda_2 T(p)^2 h(p_2) \delta(\varphi)$$
$$+ \lambda_1 T(p) \delta(\varphi) h(p_1) \int_\Omega T(q) H(\varphi) h(p_1) dq$$
$$- \lambda_2 T(p) \delta(\varphi) h(p_2) \int_\Omega T(q)(1 - H(\varphi)) h(p_2) dq, \quad (8)$$

where $T$ denotes one region from a raw image of the same size with the corresponding DEL. To solve the above function, Eq. (8), we use a similar level set method with Ref. 20 to reinitialize $\phi$ to its zero-level curve because with reinitialization, the curve can obtain a characteristic steady state (see more details in Ref. 20).

### 2.4 Iteration Detection

To observe the group behavior of a crowded population, frequent occlusions and distractions are inevitable, which will bring a big challenge to the proposed detection and segmentation methods. From the large numbers of objects as shown in Fig. 1, we find a clue that although objects in some regions are very crowded and some cannot be distinguished even by the human eye, we can still judge those objects around crowded regions. We term this clue as clue-2, which gives us the idea to first detect the outside targets. Therefore, we propose an iterative approach to solve these challenges.

When using DEL, we find objects outside the crowded regions have relatively large fit values $\gamma$ and those inside the regions have relatively small $\gamma$ (see details in Sec. 2.2). If the specified threshold $\lambda$ is too large, only some of the objects will be detected, whereas if $\lambda$ is too small, many incorrect locations are mistakenly marked as targets. In order to refine the results, we use an iterative approach to change $\lambda$ dynamically. $\lambda$ will decrease by $\tau$ and update as $\lambda - \tau$ after each loop, i.e., first locate those objects that have large fit values. Other parameters used in Secs. 2.2 and 2.3 are reinitialized as their original values. The regions detected and segmented at previous loops should be labeled as processed regions and will not be detected at the subsequent loops. Generally, we will adopt satisfying results after several times of iteration.

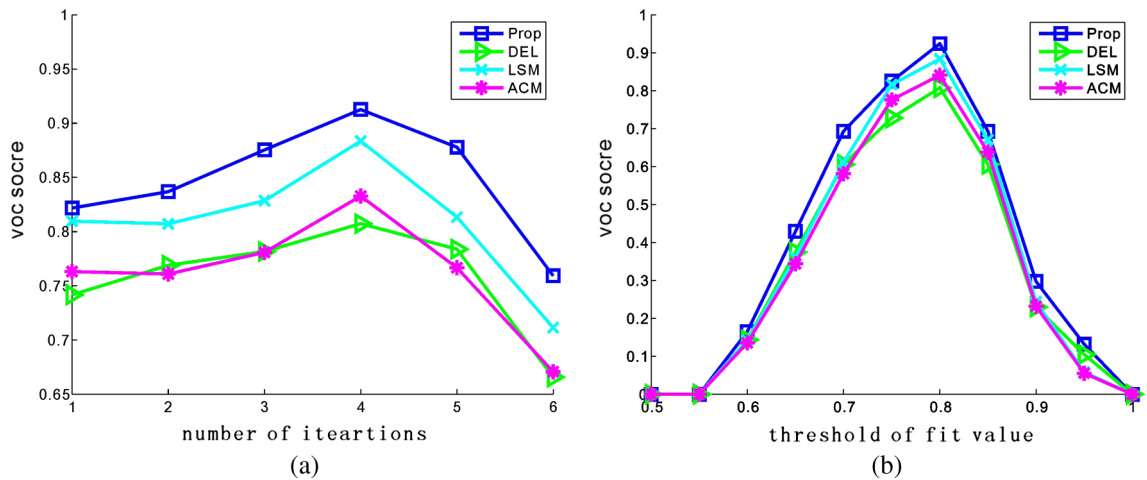

**Fig. 3** Performance of the compared methods on paramecium data set. (a) Visual object classes (VOC) score versus the number of iterations ($\lambda = 0.82$ and $\tau = 0.01$). (b) VOC score versus the threshold $\lambda$ of fit value (the number of iterations is 4). One can see that the proposed method performs best.

Figure 3 gives a detailed discussion about the iterative detection and segmentation.

Although iteration of detection and segmentation methods can significantly improve accuracy, occlusions and detection missing still occur. We will use the proposed tracking and linking methods as discussed in Sec. 3 to overcome the remaining challenges.

## 3 Data Assignment

Object tracking for large numbers of high-density populations is a very challenging task because it is very difficult to maintain the identities of visually similar targets in the presence of occlusions throughout the whole sequences to obtain complete trajectories. In this section, we do not use traditional methods to define the similarity function as discussed in Sec. 1, but propose trainable cost matrix of assignment while fully using spatial–temporal information, texture, and shape features. Then, we formulate trajectory measurement as two steps of LAP.

### 3.1 Training and Tracking

For object tracking, association cost between consecutive frames should be defined first. To the best of our knowledge, most of the traditional methods are prone to defining a similarity function by fusing spatial and texture features yet ignoring the important shape feature. However, when dealing with trajectory measurement of deformable objects, such as zebra fish and cells, shape feature provides an important clue. In addition, definition of a similarity function by fusing many features has an obvious weakness in that the cost matrix highly depends on the weight of each term and a suitable normalization method should be employed for each term separately, which is very subjective. We propose the idea of a trainable cost matrix of assignment with many spatial—temporal, texture and shape features.

RF classifier[21] has a strong ability to handle many features, which is borrowed in our trajectory measurement. To train an RF classifier, we first need to collect both positive and negative samples. In fact, only positive assignment samples need to be collected by users because for an object between consecutive frames, there is only one correct assignment, and the others are wrong. We take correct assignments as positive samples, and the wrong assignments as negative samples.

The features we used to train an RF classifier include: distance of object center, overlap of object areas, difference of object size, shape similarity, texture similarity, and direction similarity.

#### 3.1.1 Shape similarity

Ray descriptor[22] claimed to be a very effective method to measure object shape. The ray feature set contains distance difference feature $f^{[\text{diff}]}$, distance feature $f^{[\text{dist}]}$ and norm feature $f^{[\text{norm}]}$. It is important that the descriptor is the same no matter the orientation of object. This is the main reason for using this kind of shape feature.

#### 3.1.2 Texture similarity

Texture is an important visual feature in object tracking. We use the intensity histogram to denote texture feature and measure the similarity by Euclidean distance.

#### 3.1.3 Direction similarity

For the corresponding objects between consecutive frames, there will be a moderate deflection angle. We use this angle to denote their direction similarity.

After training and prediction by RF classifier, we can obtain the votes $\mathcal{V}(i, j)$ for object $i$ and $j$ in consecutive frames. Let $\mathcal{V}$ be a cost matrix, and we then transform trajectory measurement to a LAP, which can get optimal solution by the Hungarian algorithm.[23]

### 3.2 Linking

Tracking by LAP is effective when there are only a constant and small number of objects. In our tracking case, however, assignment faces large difficulties because of trajectory tracklets: (1) frequent occlusions are inevitable when handling high-density population, (2) frequent collision and extrusion lead to a certain number of missed detections, and (3) some targets entering or moving out of the field of view lead to variable assignment selection in binary matching. To obtain complete and satisfying trajectories, linking these tracklets is necessary.

Let $\mathfrak{I}_k$ be tracklet $k$. A potential complete trajectory linked by $\mathfrak{I}_i$, $\mathfrak{I}_j$, and $\mathfrak{I}_k$ (may be more) should satisfy the constraint of time continuity and similarity of object feature. In fact, we can naturally think a real trajectory $\mathcal{T}$ is the combination of several tracklets, which can be a simple equivalent for the connection of any two tracklets, e.g., $\mathcal{T} = \{\mathfrak{I}_i \sim \mathfrak{I}_j \sim \mathfrak{I}_k\} = \{\mathfrak{I}_i \sim \mathfrak{I}_j \text{ and } \mathfrak{I}_j \sim \mathfrak{I}_k\}$. Similar to object tracking in Sec. 3.1, our problem can now be reduced to a second LAP, whose goal is to find the optimal pairwise linking. After this linear assignment, a recursive approach can easily link these assigned tracklets into complete trajectories. Specifically, when there are frequent occlusions, detection misses usually occur, which will cause trajectory fragments. After we use the second LAP to link these tracklets, some dummy objects will be added in the places where detection misses occur.

## 4 Experiment

### 4.1 Data Sets

Paramecium and zebra fish are widely used as targets to study the behavior of swimming objects.[24,25] In order to evaluate the proposed method, we perform our system on the challenging real-world databases as shown in Figs. 1 and 4. These crowded objects often make current segmentation and tracking methods prone to failure because of frequent occlusions and obscure boundaries. The paramecium data set contains 110 images of size $708 \times 532$, which was provided by Prof. Tamas Vicsek[2,4] from Eotvos Lorand University in Hungary and the zebra fish data set contains 617 images of size $1024 \times 1024$, which was captured by an OPAL-1000 digital CCD camera.

To demonstrate the wide applicability of the proposed framework, we also apply our method on mitochondrion segmentation.[6] Specifically, the geometry and spatial distribution of mitochondria contain highly valuable information for scientific investigation and medical diagnosis related to neuronal functionality[26] and diseases, e.g., Parkinson's.[27] The advent of serial section scanning electronic microscope (ssSEM) has made it possible to obtain precise data of their three-dimensional (3-D) geometry information and spatial distribution. In this young field, often termed *connectomics*,[28] the task of retrieving such data from ssSEM stacks is, however, quite challenging due to variability of shape and texture of mitochondria as well as complex backgrounds such as membranes and synapse that may easily be mistaken as mitochondria even by human eyes.[29] Therefore, research to develop effective methods to segment mitochondrion from ssSEM image sequence has started to attract attention in recent years. The mitochondrion data set we used in the experiment represents a $20 \times 20 \times 4$ μm$^3$ section, corresponding to a $4096 \times 4096 \times 100$ volume (about $1.68 \times 10^9$ voxels), which was captured by *FEI Helios NanoLab 600i* FIB-SEM. The resolution of each voxel is about 5 nm in x- and y- directions and 40 nm in z-direction.

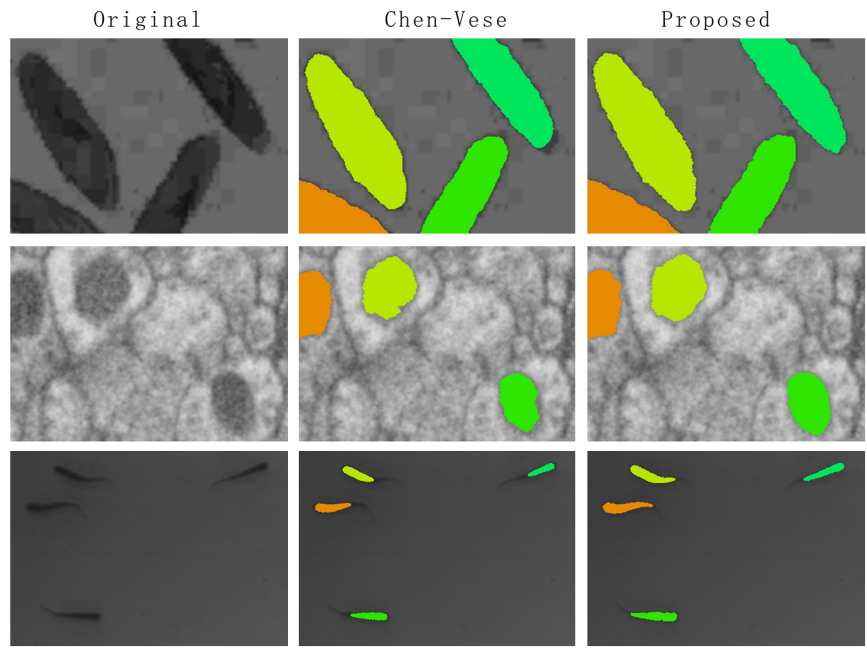

**Fig. 4** Visual segmentation comparison for paramecium (row 1), mitochondria (row 2), and zebra fish (row 3) by Chen–Vese (Ref. 19) and our method.

### 4.2 Parameters and Implementation Details

In Sec. 2.2, the scalable parameter $k = \sqrt{G_1/G_2}$ is usually specified as 0.9. The other two scalable parameters $a$ and $b$ are used as the mean length and width of testing targets. $c$ is often specified as the possible change of $a$ and $b$ experientially, which is specified as a fixed value 2 in our experiment. Generally, when employing difference of Gaussian model to fit targets in an image such as in Ref. 7, at least five parameters such as $\sigma_{x_1}$, $\sigma_{y_1}$, $\sigma_{x_2}$, $\sigma_{y_2}$ and $c$ should be specified to control the scalable evolution, however, we only need four parameters, $k$, $a$, $b$, and $c$, because of the advantages of the proved DEI, as discussed in Sec. 2.1. In general, users only need to specify $a$, $b$, and $c$ based on different data sets accordingly, and other parameters can be used as the recommended values.

For the proposed VM-ACM, $\lambda_1$, $\lambda_2$, and $\lambda_3$ are usually specified as 1 because each item in Eq. (7) is normalized. When dealing with high-density population, iterative detection one time usually cannot obtain satisfying results because of frequent occlusions, as discussed in Sec. 2.4. Therefore, an iterative approach to change fit value is necessary. Generally, we specify a relatively big threshold $\lambda$ of fit value and gradually decrease by $\tau$ after each iteration. For paramecium and zebra fish, we specify $\lambda = 0.82$ and $\tau = 0.01$, and after four iterations we could detect most of the targets. For mitochondrion, because the objects do not often touch each other, iterative detection only one time can adopt satisfying results ($\lambda = 0.8$). Figure 3 gives a detailed discussion about the number of iteration and selection of fit values. In the real experiment of other data sets, we suggest users specify $\lambda \in [0.75, 0.85]$, $\tau \in [0.01, 0.02]$, setting $a$ and $b$ as the mean length and width of the targets, fixing $c$ experientially and iterating (Secs. 2.2 and 2.3) two to four times.

For tracking, we use 100 correct assignments as positive samples and 100 incorrect assignments as negative samples. The RF classifier ensemble consists of 50 trees. Notably, there is only one correct assignment for a target between consecutive frames. Therefore, in experiment, only correct assignments need to be collected by users and negative samples are selected randomly from incorrect assignments.

### 4.3 Segmentation Evaluation and Comparison

Rand index and Jaccard index are two types of methods in the computer vision community to evaluate segmentation quality. Rand index, however, does not account for all types of topological errors,[29] and we use the defacto standard Jaccard index [Visual object classes (VOC) score] to evaluate

**Table 1** VOC score for the proposed and current state-of-the-art methods (%).

|              | Prop  | DEL   | EF    | LSM   | ACM   |
|--------------|-------|-------|-------|-------|-------|
| Paramecium   | 92.40 | 80.73 | 70.38 | 88.35 | 86.45 |
| Mitochondrion| 89.26 | 75.88 | 70.34 | 4.77  | 85.69 |
| Zebra fish   | 93.82 | 84.59 | 72.30 | 90.18 | 87.05 |

the proposed method. The Jaccard coefficient can be denoted by

$$\text{VOC} = \frac{\text{TruePos}}{\text{TruePos} + \text{FalsePos} + \text{FalseNeg}}. \quad (9)$$

We choose 30 frames from the whole image sequences, respectively, and manually detect the targets as the ground truth. Table 1 gives VOC scores of the proposed (Prop) and other state-of-the-art methods on three different databases. DEL in Table 1 denotes the proposed iterative coarse detection method without the proposed VM-ACM as discussed in Sec. 2.3. We can see that without this procedure, DEL falls behind Prop about 10%, which shows that the proposed ACM can obtain better segmentation results than the coarsely detected objects by DEL. Level set method (LSM) and ACM in Table 1 denote the VOC score of current state-of-the-art methods[30] and Chen–Vese's active contour without edge model.[19] For a fair comparison, we follow a similar line of procedure and use LSM and ACM instead of the proposed VM-ACM to obtain the fine segmentation results based on the same coarsely detected objects. From Table 1, one can see that VM-ACM performs best. After exhaustive research, we find LSM is prone to mistaking the dark blobs inside paramecium as target objects and when dealing with obscure boundaries, both LSM and ACM cannot obtain VM segmentation. EF denotes the elliptical filter method proposed by Chaudhury et al.,[7] which is similar with the proposed DEL and uses a difference of Gaussian model to detect cells. They did not, however, fully use the property of DEI stated in Sec. 2.1 and did not discuss the case of high-density population. In addition, without iterative process, EF falls behind DEL about 10%.

Compared with the ground truth, the proposed method can obtain 92.40%, 89.26%, and 93.82% scores for paramecium, mitochondria, and zebra fish school, respectively. The errors of paramecium and zebra fish school mainly come from frequent occlusions, which will be improved by the proposed tracking and linking method, as discussed in Secs. 3.1 and 3.2. The errors of mitochondria mainly come from the connected membranes and synapses, which can be discarded with spatial information and 3-D texture. We will not discuss the detailed procedure about mitochondrion in this paper.

Figure 3 shows the VOC scores versus the number of iterations and threshold $\lambda$ of fit value. One can see that, for a high-density paramecium data set, one iteration could not adopt satisfying segmentation results. This result verifies clue-2 as discussed in Sec. 2.4. However, it does not mean that the VOC score will always increase as the number of iterations increases, because $\lambda$ will decrease by $\tau$ after each iteration and a large number of iterations will indicate a small fit value, which are prone to causing many detection errors. In the experiment of other data sets with high-density targets, we find that the system can obtain satisfying results after two to four iterations.

To further declare the superiority of the proposed VM-ACM against Chen–Vese's active contour without edge model,[19] we give the visual segmentation comparison on real-world databases of paramecium, mitochondria, and zebra fish as illustrated in Fig. 4. Decided by the energy function Eq. (6),[19] this is prone to decreasing the area and length of active contour. Therefore, when dealing with long and thin tails of zebra fish, their method, usually, could not obtain satisfying segmentation results. In addition, when objects have similar intensity and texture with background, Chan–Vese[19] may mistake the obscure boundaries as background such as the zebra fish segmentation in Fig. 4. In contrast, the proposed VM-ACM can obtain better results.

### 4.4 Tracking Evaluation and Comparison

The multiple object tracking accuracy (MOTA) and multiple object tracking precision (MOTP) error measure are commonly used by the target tracking community.[31] They are computed as

$$\text{MOTA} = 1 - \frac{\sum_t (m_t + fp_t + mme_t)}{\sum_t g_t} \quad (10)$$

and

$$\text{MOTP} = \frac{\sum_{i,t} d_t^i}{\sum_t g_t}, \quad (11)$$

where $m_t$, $fp_t$ and $mme_t$ in MOTA denote the numbers of misses, false positives, and mismatches for time $t$, respectively. $d_t^i$ in MOTP denotes the Euclidean distance between object $o_i$ and its corresponding hypothesis. $g_t$ in MOTA and MOTP denotes the total number of objects in frame $t$.

Let T2LAP denote the proposed tracking method, which contains trainable assignment cost matrix and two steps of LAP. To evaluate the proposed linking process, we use TLAP to denote the trajectory results without the second LAP as discussed in Sec. 3.2. Similarly, to evaluate the superiority of LAP, we use nearest neighbor method instead of LAP to track and link objects. We term this method as NN. For a fair comparison, we use the same trainable cost matrix as the input of NN. From the quantitative comparison in Table 2, we can see that the MOTA score of T2LAP outperforms TLAP more than 6%. The reason is that when dealing with frequent overlap and occlusions, detection misses usually occur, which will lead to uncompleted trajectory fragments. However, TLAP without a linking procedure could not retrieve complete trajectories. The MOTA score of T2LAP outperforms NN about 10% because when dealing with large numbers of crowded objects, individuals may have similar visual features with their neighbors, and NN is prone to finding the wrong matching. The mismatch ratio in Eq. (10) is used to count all occurrences where the tracking hypothesis for an object changed compared to previous frames,[31] which could happen when two or more objects are swapped as they pass close to each other or when an object tracker is reinitialized with a different track ID after it was previously lost because of occlusion or detection miss. In Fig. 5, we can see that TLAP and NN have larger mismatch ratios (10.4% and 20.65%, respectively) than the proposed T2LAP (6.03%). With the second LAP, many tracklets caused by occlusions or detection misses will be linked to complete trajectories and the mismatch ratio increases by 4.37% and the MOTA ratio increase by 6.67% for paramecium data set as shown in Fig. 5. Because the proposed linker will add some dummy objects in the positions where occlusions and detection misses occur, the MOTP score of T2LAP will also increase by 3.58 pixels. Specifically, the MOTP scores of T2LAP and TLAP are 8.97 and 5.39 pixels. The MOTP of TLAP is mainly caused by detection and segmentation error as discussed in Secs. 2.2 and 2.3. The dummy object is added at the average position of its previous and next objects as illustrated in Fig. 6.

**Table 2** MOTA comparison of the proposed method and current state-of-the-art methods (%).

|            | T2LAP | TLAP  | T2LAP-S | NN    | K2LAP | IMM   |
|------------|-------|-------|---------|-------|-------|-------|
| Paramecium | 89.39 | 82.69 | 87.68   | 79.35 | 80.91 | 84.32 |
| Zebra fish | 95.37 | 88.42 | 92.51   | 83.96 | 86.53 | 89.77 |

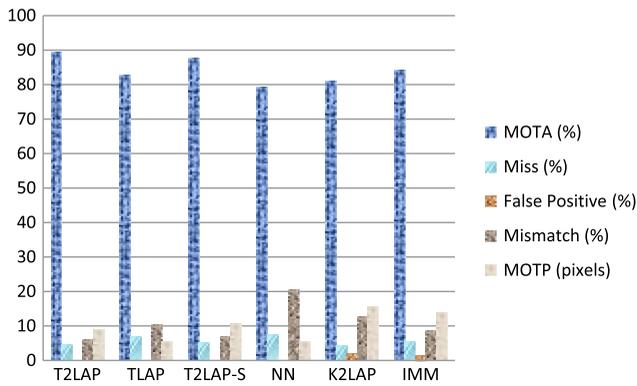

**Fig. 5** MOTA and MOTP measurement of paramecium data set for the proposed and current state-of-the-art tracking methods.

K2LAP (Ref. 5) and interacting multiple model (IMM) (Ref. 8) are considered to be current state-of-the-art methods in trajectory measurement, which can be used to track large numbers of objects. Wu et al.[5] used the spatial distance as the cost matrix and employed a Kalman filter to predict positions of objects in the next frame based on their coordinates and velocities. However, the Kalman filter used for dynamic filtering is bound to use one dynamics model only, which can be problematic as the dynamics of cells may vary frequently with time. Li et al.[8] adopted the IMM filter, which allowed multiple dynamics models in parallel, and was shown to be more biologically relevant than Kalman filter. We use coordinate of center, velocity, and direction of body as the state variables of their dynamic models to predict the next hypothesis for each object. As shown in Table 2 and Fig. 5, their methods are prone to failing when dealing with crowded objects where there are frequent occlusions. In addition, K2LAP and IMM have high MOTP scores, 15.72 and 13.86 pixels, respectively, which indicate that dynamic models like Kalman filter and particle filter are not suitable for predicting irregular movement of objects such as paramecium, as illustrated in Fig. 7.

In addition, to the best of our knowledge, when dealing with tracking, many researchers, Wu et al.[5] and Li et al.,[8] would like to use spatial and texture features, yet they rarely employed shape features. In this paper, we use Ray feature[22] to describe the shape of different deformable objects, and we propose the idea of using trainable cost matrix of assignment with many spatial, texture, and shape features. To evaluate the benefit of imported shape features, let T2LAP-S denote the method only with spatial and texture features. From Table 2 and Fig. 5, we can see that with ray descriptor shape feature, the MOTA of T2LAP outperforms T2LAP-S about 2%.

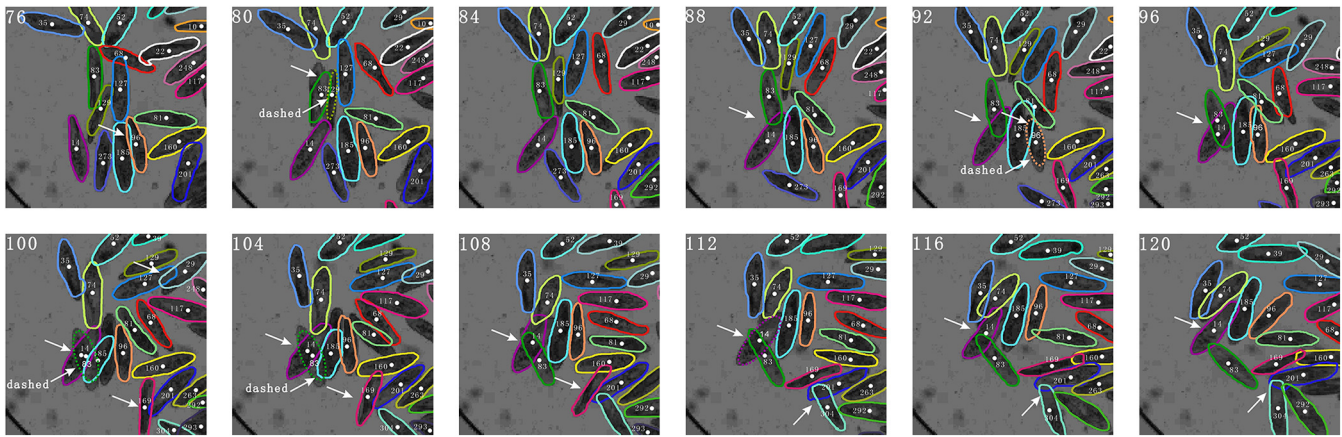

**Fig. 6** Tracking high-density paramecium population through occlusions (zooming in for more details). The enlarged parts of frame 76 to 120 are shown here. White arrows indicate the occurrence of occlusions. One can see that the identities of paramecium are correctly maintained although the occlusions last for several frames. The detection misses are represented by dummy objects (dashed ellipses).

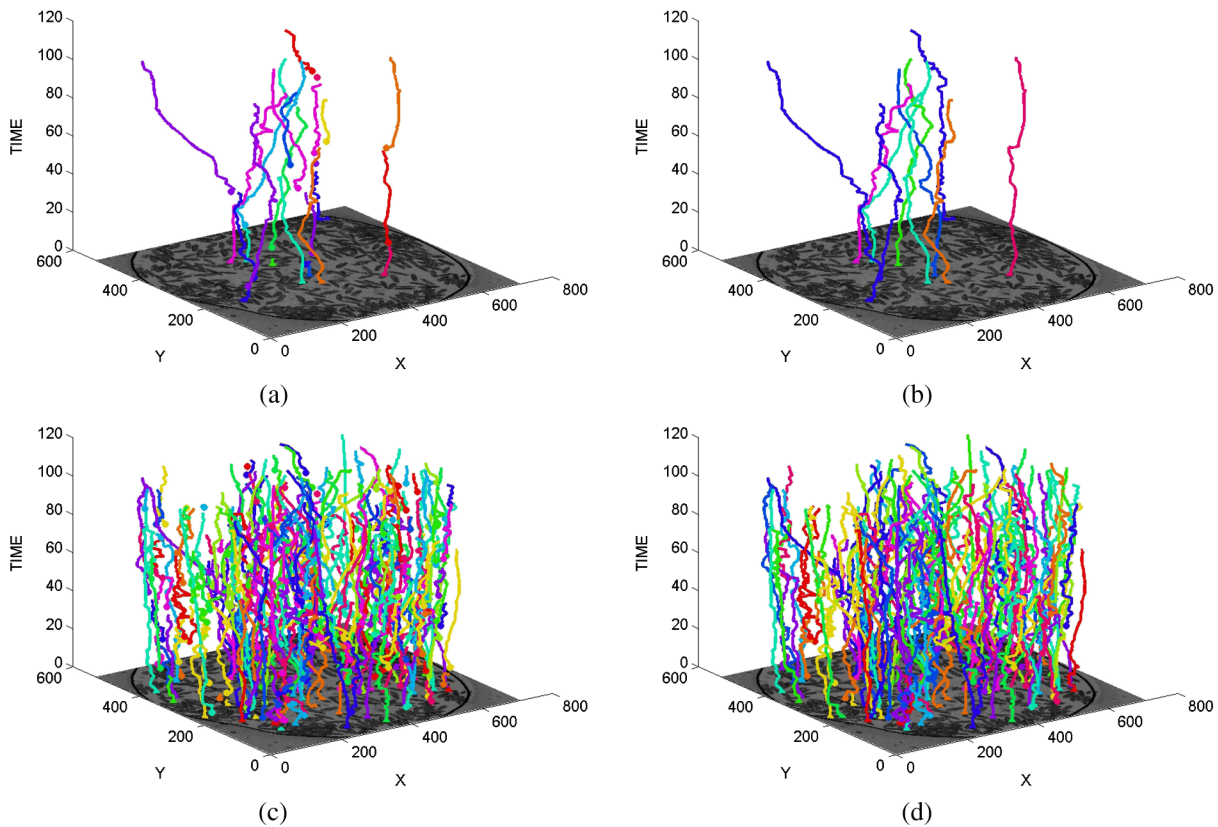

**Fig. 7** Spatial–temporal plots of tracklets obtained by the first LAP (tracking) and complete trajectories adopted by the second LAP (linking). For clarity, only partial results are shown here. Because of occlusions and detection misses, many tracklets will be obtained by the first LAP as illustrated in (a) and (c). After performing tracklet linking with the second LAP, correct and complete trajectories are acquired as shown in (b) and (d). More tracking and linking results are shown in (c) and (d). Trajectory beginning point is marked by a round dot.

## 5 Discussions and Conclusions

We present an automatic and effective framework for measuring trajectories of large numbers of crowded oval-shaped objects. We first prove an interesting DEI, namely that the zero and minimum level sets of arbitrary DoG functions of the same center, phase, and maximum are circumscribed and inscribed ellipses of a rectangle. This inference reveals an important property of DoG model and can reduce the dimension of DoG model without any loss of precision. The property can be used in many other applications and can reduce their computation cost such as SIFT (Ref. [17]), band-pass filter, Laplacian of Gaussian, Mexican Hat function, etc.

For object detection, we first use DEL based on the proved DEI to detect the coarse positions of targets and then utilize the proposed VM-ACM to get the fine segmentation results because accurate segmentation results and object shape are very important to biologists.[8,18] In addition, feature extraction based on segmentation results rather than detection results is more precise, which will be used in object tracking and linking steps as discussed in Sec. 3. To demonstrate the wide applications of the proposed segmentation method, we perform our algorithm on several different data sets including paramecium, zebra fish, and mitochondrion as illustrated in Fig. 4. We give detailed analysis and comparison about the proposed method and the state-of-the-art methods. Experimental results show that the proposed method performs best. In the future, we will try to define more a complicated shape model to fit objects with other irregular shapes.

We find two useful clues for the detection of high-density oval-shaped population: clue-1 in Sec. 2.2 and clue-2 in Sec. 2.4. Specifically, clue-1 indicates the use of DEL to detect the objects and clue-2 indicates the use of iterative approach to obtain higher detection accuracy. We think that these two clues give a possible approach to overcome the challenges in detection and tracking of crowded objects.

For the definition of assignment cost, in addition to the use of spatial–temporal and texture features,[8] we also use shape feature to define the similarity between two objects. Experimental results show that shape feature provides an important information for the trajectory measurement of deformable objects as illustrated in Fig. 5. In addition, we do not use a traditional method to define an energy function with fusing many features, but propose the idea of a trainable cost matrix with spatial–temporal, texture, and shape features. The weakness of traditional methods are discussed in Sec. 1. The proposed idea of a trainable cost matrix can easily be used in other applications which need to define the similarity between two objects.

For data association, we formulate trajectory measurement as two steps of LAP. The first LAP can obtain many trajectories of objects by employing Hungarian algorithm. Because of frequent occlusions and detection misses, many tracklets (trajectory segments) occur. Then, we use the second LAP to link tracklets and adopt complete trajectories. In addition, the second LAP can suggest adding some dummy objects where tracklets occur. These two LAPs work collaboratively to reliably track large numbers of a crowded population. Experimental results show that the proposed tracking method outperforms the state-of-the-art methods. Our tracking and linking method with two steps of LAP can also be used in other applications with trajectory measurement.

The limitation of the proposed approach is that it requires a lot of time to obtain the accurate segmentation of each object, which cannot be used in real-time applications at its current stage. Specifically, it will cost about 819.7s to segment 300 paramecium in an image of size $708 \times 532$ with four iterations (Intel i7-2600). However, the proposed framework is very easy to perform concurrently on multicores machines. The future work is to develop a more efficient system based on graphics processing unit platform. In addition, the work presented in this paper addresses oval-shaped objects, such as cells, zebra fish, and human heads. In the future, we plan to address other objects with irregular shapes. As a subject of future work, each of the components of the proposed framework may be improved or replaced to suit different problems. For example, while we use DEL to obtain the coarse position of each object, it would be possible to replace this with other detectors. The proposed detection, segmentation, and tracking methods can be individually used in other applications such as flocking birds and flying drosophila. In addition, our framework can also potentially be used for 3-D tracking task in multiview applications.

*Acknowledgments*

The research work presented in this paper is supported by National Natural Science Foundation of China, Grant No. 61175036. We would like to thank Professor Tamas Vicsek from Eotvos Lorand University in Hungarian for providing paramecium data set in our experiments. For mitochondria segmentation, we would like to thank Bangyu Zhou, Guangxia Wang and Ke Zhang in the State Key Laboratory of Neuroscience, Chinese Academy of Sciences for providing electronic microscopy data set of the $\alpha$-lobe of drosophila. In addition, we would like to thank Jinming Guo in Shanghai University and Shuohong Wang in Fudan University for their helpful discussions and data analysis. We would like to thank Jun Ma and Krystle Thai for their help to revise the manuscript. We also acknowledge the anonymous reviewers for their valuable comments.

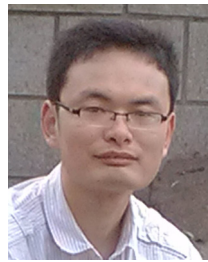

**Hui Li** received his BS degree from School of Computer Science and Engineering at Shanghai University in 2011. He is now a candidate Master student in School of Computer Science, Fudan University, Shanghai, China. His current research interests include 3D reconstruction, object tracking, biomedical image processing and machine learning.

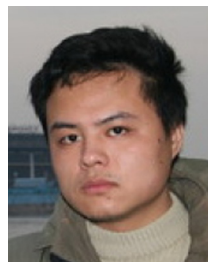

**Ye Liu** received his BS degree from the Department of Computer Science and Technology at Tongji University, Shanghai, China, in 2007. He is now a PhD student in School of Computer Science, Fudan University, Shanghai, China. His current research interests include 3D computer vision and pattern recognition, with a special attention to 3D object detection and tracking.

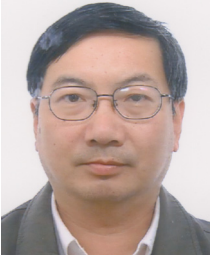

**Yan Qiu Chen** received his PhD degree from Southampton University, United Kingdom in 1995, and his MEng and BEng degrees from Tongji University, Shanghai, China in 1988 and 1985 respectively. He is currently a full professor with School of Computer Science of Fudan University, Shanghai, China, and is a member of Fudan University Academic Committee, and chairman of Computer Science School Academic Committee. He had been chairman of Department of Communication Science and Engineering from 2004 through 2007, and associate chairman of Department of Computer Science and Engineering from 2002 through 2004. He was an assistant professor with School of Electrical and Electronic Engineering of Nanyang Technological University, Singapore from 1996 through 2001, and was a postdoctoral research fellow with Glamorgan University, UK in 1995.